\documentclass[11pt]{article}

\usepackage[final]{acl}

\usepackage{times}
\usepackage{latexsym}

\usepackage[T1]{fontenc}
\usepackage[utf8]{inputenc}
\usepackage{makecell}
\usepackage{microtype}

\usepackage{inconsolata}

\usepackage{graphicx}

\usepackage{enumitem}
\usepackage{multirow}
\usepackage{wrapfig}
\usepackage{colortbl}
\usepackage[normalem]{ulem}
\usepackage{comment}
\usepackage{microtype}
\usepackage{booktabs}

\usepackage{amssymb}

\setlist[itemize]{align=parleft,left=0pt,topsep=1mm,itemsep=0mm,parsep=1mm}

\definecolor{azure(colorwheel)}{rgb}{0.0, 0.5, 1.0}
\definecolor{nicegreen}{rgb}{0.0, 0.7, 0.1}
\definecolor{yw}{rgb}{0.01176, 0.5490, 0.5490}
\definecolor{ashblue}{rgb}{0.36, 0.54, 0.66}
\definecolor{ashgrey}{rgb}{0.7, 0.75, 0.71}
\definecolor{applegreen}{rgb}{0.55, 0.71, 0.0}
\definecolor{greenyellow}{rgb}{0.68, 1.0, 0.18}
\definecolor{junebud}{rgb}{0.74, 0.85, 0.34}
\definecolor{jh}{rgb}{0.3, 0.73, 0.09}
\definecolor{ywg}{rgb}{0.9960, 0.8984, 0.5859}
\definecolor{jy}{rgb}{0.58, 0, 0.827}
\definecolor{cornellred}{rgb}{0.7, 0.11, 0.11}
\definecolor{darkcyan}{rgb}{0.0, 0.55, 0.55}
\definecolor{CuGray}{gray}{0.9}
\definecolor{airforceblue}{rgb}{0.36, 0.54, 0.66}
\definecolor{rev}{rgb}{0.784, 0.003, 0.313}
\definecolor{pink}{cmyk}{0, 0.7808, 0.4429, 0.1412}
\definecolor{amethyst}{rgb}{0.6, 0.4, 0.8}
\definecolor{black}{rgb}{0.0, 0.0, 0.0}
\definecolor{tb3_yellow}{rgb}{0.996, 1.0, 0.6}
\definecolor{tb3_orange}{rgb}{0.980, 0.8, 0.604}
\definecolor{tb3_red}{rgb}{0.972, 0.6, 0.6}
\definecolor{dimgray}{rgb}{0.41, 0.41, 0.41}
\definecolor{brickred}{rgb}{0.8, 0.25, 0.33}
\definecolor{bleudefrance}{rgb}{0.19, 0.55, 0.91}
\definecolor{blue(ncs)}{rgb}{0.265, 0.445, 0.765}
\definecolor{green(ncs)}{rgb}{0.0, 0.62, 0.42}
\newcolumntype{g}{>{\columncolor{CuGray}}c}
\newcolumntype{z}{>{\columncolor{CuGray}}l}

\renewcommand{\paragraph}[1]{\vspace{0.5mm}\noindent\textbf{#1.}\,\,}

\newcommand{\mok}[1]{\textcolor{black}{#1}}

\usepackage{xspace}

\makeatletter
\def\@fnsymbol#1{\ensuremath{\ifcase#1\or *\or \dagger\or \ddagger\or
   \mathsection\or \mathparagraph\or \|\or **\or \dagger\dagger
   \or \ddagger\ddagger \else\@ctrerr\fi}}
\makeatother

\def\onedot{.\@\xspace}
\def\eg{\emph{e.g}\onedot} 
\def\ie{\emph{i.e}\onedot}

\newcommand{\Sref}[1]{Sec.~\ref{#1}}

\newcommand{\Fref}[1]{Fig.~\ref{#1}}
\newcommand{\Tref}[1]{Table~\ref{#1}}

\newcommand{\be}{\begin{eqnarray}}
\newcommand{\ee}{\end{eqnarray}}
\newcommand{\bee}{\begin{eqnarray*}}
\newcommand{\eee}{\end{eqnarray*}}

\newcommand{\matrixb}{\left[ \begin{array}}
\newcommand{\matrixe}{\end{array} \right]}

\usepackage{mathtools}

\DeclarePairedDelimiterX{\inp}[2]{\langle}{\rangle}{#1, #2}

\title{SMILE-Next: Teaching Large Language Models \\to Detect, Classify, and Reason about Laughter}

\author{
 \textbf{Lee Jung-Mok\textsuperscript{1}}  \ \
 \textbf{Kim Sung-Bin\textsuperscript{2}}  \ \
 \textbf{Joohyun Chang\textsuperscript{3}} \ \
 \textbf{Lee Hyun\textsuperscript{2}}      \ \
 \textbf{Tae-Hyun Oh\textsuperscript{3}}  
 \\
 \textsuperscript{1}School of EE, KAIST,  
 \textsuperscript{2}Dept. of EE, POSTECH,
 \\
 \textsuperscript{3}School of Computing, KAIST
\\
 \small{
   \textbf{Correspondence:} \href{mailto:taehyun.oh@kaist.ac.kr}{taehyun.oh@kaist.ac.kr}
 }
}

\begin{document}
\maketitle
\begin{abstract}
Laughter is a complex social signal that conveys communicative intent beyond amusement. While prior work has focused on isolated laughter analysis tasks, a comprehensive understanding of laughter in real-world scenarios remains underexplored. Therefore, we introduce SMILE-Next, a dataset for real-world laughter understanding with multimodal textual representations and question–answer annotations across three tasks: laughter detection, laughter type classification, and laughter reasoning. 
Building upon SMILE-Next, we aim to develop a laughter-specialized large language model capable of nuanced understanding of laughter in real-world contexts. To this end, we propose two key components: laughter-specific Self-Instruct and the Mixture-of-Laugh-Experts (MoLE) framework.
Laughter-specific Self-Instruct enhances generalization across tasks and domains by automatically synthesizing diverse laughter-centric instructions.
MoLE introduces a task-adaptive expert routing mechanism that dynamically selects specialized experts tailored to each laughter-related task, improving task-specific performance and efficiency.
Experimental results show that the combination of our proposed components substantially outperforms multimodal LLM baselines, advancing robust real-world laughter understanding. Project page is at: \url{https://mok0102.github.io/smile-next/}.

\end{abstract}

\section{Introduction}
\label{sec:introduction}
Laughter is one of the most fundamental and universal forms of human expression in social interactions. Beyond a simple reaction, people often laugh not only because something is funny but also to build intimacy~\citep{stauffer1999let}, capture attention~\citep{wanzer2010explanation}, reinforce rapport, or ease tension~\citep{scott2014social}. As such, laughter embodies nuances that spoken language alone often cannot convey, and we naturally interpret the intentions behind laughter to sustain social interaction.

Following recent advancements in artificial social intelligence~\citep{bainbridge1994artificial, dautenhahn2007socially, williams2022supporting}, an emerging line of work has explored building systems capable of understanding laughter or humor~\citep{bertero-fung-2016-deep,hasan-etal-2019-ur,alnajjar-etal-2022-laugh,hessel-etal-2023-androids,ko2023can,xie2024funqa,smile24hyun}. However, these approaches primarily focus on narrowly defined tasks, such as detecting humor~\citep{bertero-fung-2016-deep,hasan-etal-2019-ur}, estimating humor intensity~\citep{alnajjar-etal-2022-laugh}, reasoning about humor in cartoons~\citep{hessel-etal-2023-androids} or in general textual and multimodal content~\citep{ko2023can,xie2024funqa}, and analyzing laughter in TV shows~\citep{smile24hyun}. As a result, they are limited in capturing laughter naturally occurring in real human–human interactions and in providing a broad, context-aware understanding of laughter, essential for enabling richer human–machine and machine–machine social interactions.

\begin{figure*}[t]
    \centering
    \includegraphics[width=1\textwidth]{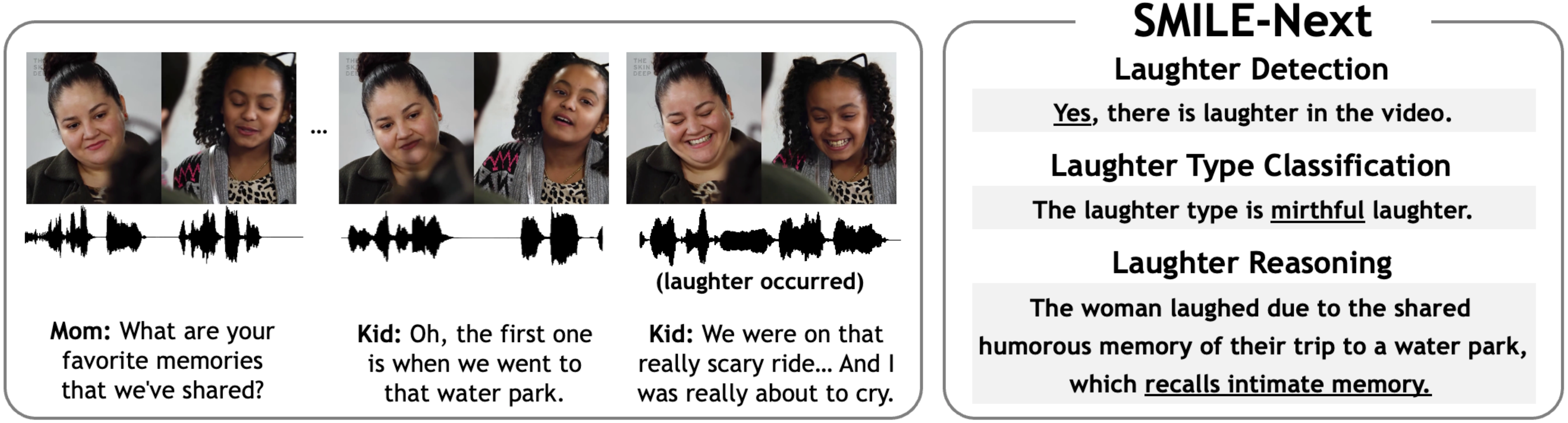}
    \vspace{-5mm}
    \caption{\textbf{Task overview of SMILE-Next.} We present SMILE-Next, a comprehensive dataset for laughter understanding. The dataset includes three fundamental tasks for understanding laughter in social interactions: laughter detection, laughter type classification, and laughter reasoning.}
    \label{fig:task_teaser}
    \vspace{-4mm}
\end{figure*}

To address this challenge, we curate \textbf{SMILE-Next}, which aims to extend machine capabilities in generalized laughter reasoning within human interactions. SMILE-Next consists of videos paired with question–answer annotations designed for comprehensive laughter understanding. The dataset encompasses three tasks:
(1) \textit{Laughter detection}, determining whether a given video is intended to induce laughter;
(2) \textit{Laughter type classification}, categorizing the type of laughter; and
(3) \textit{Laughter reasoning}, providing free-form textual explanations for why laughter occurred.
Given the diverse nature of laughter, our dataset covers a broad range of human interactions, including talk shows, sitcoms, and dyadic conversations, as illustrated in \Fref{fig:task_teaser}. Moreover, recognizing that curated data cannot cover all scenarios, we introduce a self-instructional data generation framework to overcome limitations in coverage and scale.

Building on this dataset, we develop a \textbf{laughter expert Large Language Model (LLM)}, designed to identify and interpret laughter in social interaction. Because laughter arises from complex factors including subjectivity~\citep{warren2021makes}, contextual knowledge~\citep{nijholt2017humor}, and multimodal cues, such as facial expressions and prosody~\citep{hasan-etal-2019-ur}, it is challenging for a model to interpret these entangled signals directly from video. To enable clearer interpretation by disentangling these factors, we convert multimodal signals (video and speech) into textual representations following \citet{smile24hyun}. Specifically, we employ off-the-shelf models to translate video and speech attributes into textual descriptions, and then leverage the reasoning capabilities of LLMs to identify, understand, and explain the causes of laughter. 
Furthermore, we propose a LoRA-based Mixture-of-Laugh-Experts (MoLE) framework within LLM training, which learns task-specific experts to flexibly handle and synergize multiple laughter-related tasks, including detection, type classification, and reasoning, within a unified framework.

Our experimental results show that integrating LLM reasoning capabilities with multimodal textual representations enables effective identification and interpretation of laughter in human interactions, including the ability to generate plausible explanations. Moreover, incorporating self-instruction and MoLE further enhances the model’s reasoning capacity. Through our analyses, we demonstrate that textualizing multimodal information is essential for robust laughter understanding. Our main contributions are summarized as follows:
\begin{itemize}
\item We introduce \textbf{SMILE-Next}, a comprehensive dataset containing videos, textual representations, and QAs covering diverse laughter scenarios.
\item We propose a \textbf{laughter-reasoning expert LLM} that demonstrates strong capabilities in laughter detection, laughter type classification, and laughter reasoning.
\item We show that \textbf{Laugh-tailored Self-Instruction} and \textbf{Mixture-of-Laugh-Experts (MoLE)} are effective strategies for enhancing LLMs’ capacity for laughter reasoning.
\end{itemize}

\section{Related Work}
\label{sec:related_work}
% \paragraph{Understanding laughter through models}
\paragraph{Understanding laughter in videos}
Laughter is a socially meaningful behavior closely tied to humor, sarcasm, and interpersonal bonding~\citep{scott2014social}. Most approaches model laughter or humor as signals to be detected or quantified, focusing on classification or intensity prediction using textual and multimodal cues~\citep{annamoradnejad2020colbert, weller-seppi-2020-rjokes, bertero-fung-2016-deep, alnajjar-etal-2022-laugh, castro-etal-2019-towards, hasan-etal-2019-ur, ray-etal-2022-multimodal}. These methods do not explicitly model why laughter occurs.
Recent studies have begun to explore explanation-oriented perspectives on humor and laughter~\citep{PaLM, hessel-etal-2023-androids, ko2023can, xie2024funqa}, but they remain largely domain-specific and focus on constrained or scripted scenarios. 
While \citet{smile24hyun} introduced an initial benchmark for laughter reasoning in real-world videos, it remains limited in task scope and interaction diversity. Our work extends this line of research by formulating laughter understanding as a multi-faceted problem, encompassing laughter detection, laughter type classification, and laughter reasoning across diverse human interaction scenarios.

\paragraph{Textualized approaches for video understanding}
Multimodal video-language understanding has traditionally relied on feature-based fusion, projecting visual, audio, and textual inputs into a shared latent space for downstream reasoning~\citep{lu2019vilbert, li2019visualbert, lei-etal-2018-tvqa}. 
While effective for recognition-centric tasks, recent analyses suggest that such implicit multimodal representations are less suitable for explanation-oriented reasoning, as causal, social, and contextual cues often remain entangled in latent feature spaces~\citep{buch2022revisiting, xie2024funqa, ko2023can}.

Recent studies instead explore textualization as an alternative interface for multimodal reasoning, converting visual and auditory signals into natural language descriptions that are directly consumable by large language models~\citep{socratic, wang2022language, hasan2023textmi, richet2024textualized, smile24hyun}. 
By representing multimodal information in a symbolic and interpretable form, textualization enables LLMs to better leverage their pretrained commonsense, social, and causal reasoning capabilities, and has shown particular effectiveness in reasoning-heavy tasks involving affective states and interpersonal dynamics.
Motivated by these observations, we adopt textualization as a key design choice for laughter understanding, where multiple implicit social and affective factors jointly give rise to laughter and require explicit reasoning.

\paragraph{Mixture-of-Experts in LLM}
Mixture-of-Experts (MoE) architectures have been extensively studied as a scalable framework for conditional computation and parameter specialization, and have been widely adopted in large language models to support efficient scaling and diverse task handling~\citep{shazeer2017, lepikhin2021gshard, fedus2022switch, du2022glam, shen2024mixtureofexperts}.
Beyond scalability, MoE has also been explored in instruction-tuned and multi-task LLMs as a flexible mechanism for accommodating heterogeneous task requirements within a single model~\citep{zadouri2024pushing, gou2023mixture,dou2023loramoe}.
In line with these advancements, we incorporate a parameter-efficient, LoRA-based MoE design into our laughter-understanding LLM to handle multiple laughter-related sub-tasks within a unified framework, including laughter  detection, type classification, and reasoning.

\section{SMILE-Next: Enhanced Multimodal Laughter Understanding Dataset}
\label{sec:dataset}
Our goal is to extend machine capabilities for understanding laughter in human interactions. Because laughter occurs in diverse forms across a wide range of social scenarios, analyzing and interpreting it cannot be reduced to a single, isolated task. To this end, we introduce SMILE-Next, a dataset designed to support more general and comprehensive laughter understanding in diverse multimodal contexts. SMILE-Next covers a broad range of interaction scenarios and defines multiple complementary tasks that jointly evaluate and facilitate general laughter understanding. To fully understand laughter in a given situation, a system must determine whether laughter occurs, identify the type of laughter, and reason about its underlying cause. Accordingly, our dataset captures these steps through three tasks in diverse multimodal scenarios: laughter detection, laughter type classification, and laughter reasoning, as shown in \Fref{fig:task_teaser}.

\subsection{Task Definition}
We provide detailed descriptions of the three proposed tasks below.

\paragraph{Laughter detection}%
Laughter detection is a binary classification task that aims to determine whether a given video clip induces laughter. Prior work on humor or sarcasm detection, often limited to specific scenarios, such as TED talks or sitcoms, primarily relies on audio cues to identify laughter~\citep{hasan-etal-2019-ur,castro-etal-2019-towards}. In contrast, our dataset encompasses a broader range of scenarios, including humorous, sarcastic, and awkward laughter across more diverse interaction settings, such as TV shows and dyadic conversations. As a result, effective laughter detection in SMILE-Next requires contextual reasoning over multimodal cues, including facial expressions, scene context, video captions, and acoustic signals, making the task substantially more challenging.

\paragraph{Laughter type classification}%
Laughter type classification is a multi-class classification task that identifies the type of laughter in a video. Inspired by the taxonomy proposed by 
% Tanaka et al.~
\citet{TANAKA2014314}, we classify laughter into three high-level categories: \textit{mirthful}, \textit{polite}, and \textit{satirical}.
\textit{Mirthful} laughter refers to spontaneous and joyful laughter, often triggered by humorous content or shared amusement.
\textit{Polite} laughter denotes socially motivated, non-spontaneous laughter, typically used to maintain conversational flow or express minimal agreement in formal or semi-formal interactions.
\textit{Satirical} laughter includes sarcastic, awkward, or mocking expressions, often conveying discomfort, disbelief, or social distance.

\paragraph{Laughter reasoning}% 
Laughter reasoning requires the model to infer and explain the underlying cause of laughter in a given video. Unlike the previous tasks, we formulate laughter reasoning as a free-form text generation problem, allowing models to produce natural language explanations for why a particular situation elicited laughter. This task builds upon the laughter reasoning task introduced in SMILE~\citep{smile24hyun}, which focuses primarily on humor-driven laughter in TED talks and sitcoms. In SMILE-Next, we substantially broaden the scope of laughter reasoning to include a wider range of interaction contexts, such as interviews, movies, and dyadic conversations.

\subsection{Task Formulation}
We formulate all tasks as question-answering problems, where the model generates answers given a video clip and a task-specific question prompt. Since laughter arises from multiple multimodal cues that are inherently entangled in raw video, directly interpreting laughter from video alone is challenging. Following \citet{smile24hyun}, we therefore convert videos into textualized representations of multimodal cues to facilitate disentanglement.
Formally, the task formulation is defined as $\hat{y} = D(\mathcal{Q}, \ \left\{ t_1,\  t_2, ..., \  t_k \right\})$, where $D$ denotes the LLM, $\mathcal{Q}$ is the question prompt, $\hat{y}$ is the generated answer, and ${t_1, \ldots, t_k}$ are multimodal textual representations extracted from the video $v$. This formulation enables the model to process disentangled multimodal information while fully leveraging the reasoning and comprehension capabilities of LLMs through textual representations. 

\subsection{Dataset Construction}

\begin{table}[t]
  \centering
  \resizebox{1\linewidth}{!}{
    \begin{tabular}{lcccc}
    \toprule
    \textbf{Task} & \textbf{Train} & \textbf{Val} & \textbf{Test} & \textbf{Total} \\
    \midrule
    Laughter Detection & 1565 & 460 & 359 & 2384 \\
    Laughter Type Classification & 1636 & 207 & 114 & 1957 \\
    Laughter Reasoning & 1565 & 292 & 188 & 2045 \\
    \midrule
    \textbf{Total} & \textbf{4766} & \textbf{959} & \textbf{661} & \textbf{6386} \\
    \bottomrule
    \end{tabular}
  }
  \caption{\textbf{Data statistics of SMILE-Next.} We report the number of samples for laughter-related task, along with the corresponding train, validation, and test splits.} 
  \vspace{-3mm}
  \label{tab:smilenext_statistics}
\end{table}

We design a dataset construction pipeline to collect videos and annotations for laughter understanding tasks. The pipeline integrates both automated and human-in-the-loop processes to ensure scalability while maintaining high annotation quality. It consists of four stages: video collection, multimodal textual extraction, pseudo-label annotation, and human verification. 

\paragraph{Laughter video collection}%
We collect source video clips containing laughter from diverse domains to capture a wide range of social contexts. Our sources include \textit{TED Talks} and \textit{sitcoms}, which primarily feature audience laughter elicited by humorous or satirical content. In addition, we collect in-the-wild videos from \textit{YouTube}, including dyadic conversations, talk shows, and movies\footnote{Video sources include YouTube channels, such as The Skin Deep, MovieClips, and talk shows hosted by Jimmy Fallon and Conan O’Brien.}. These in-the-wild videos cover more intimate conversational settings and include laughter that reflects relaxed, affiliative, or socially bonding reactions beyond explicit humor. After collecting the source videos, we use an off-the-shelf laughter detection model~\cite{gillick2021robust} to automatically identify video clips containing laughter. For each selected video, we further segment the clip into shorter units based on utterance boundaries, resulting in multiple laughter-related segments per source clip.

\paragraph{Textual multimodal cue extraction}%
Each video segment contains diverse multimodal cues, including visual cues (\eg, facial expressions and movements), acoustic cues (\eg, tone and pitch), and semantic cues (\eg, conversational content). To disentangle these signals, we convert them into structured textual representations. Specifically, we employ a set of expert models to extract utterances, acoustic features, facial action units, and video captions, all represented in text form.
In addition, we extract relationship information between speakers. For in-the-wild videos, we infer relationships from YouTube metadata and video descriptions. For talk shows, we assign roles such as ``host and guest,'' while for \textit{TED Talks} and \textit{sitcoms}, we label interactions as ``speaker and audience.'' 
Detailed descriptions of the multimodal extraction process are provided in the Appendix~\ref{par:appendix_text_multimodal}.

\paragraph{Pseudo-label annotation}%
Laughter reasoning is a free-form generation task and is challenging for human annotators to perform directly at scale. To address this, we generate pseudo-labels using expert large language models (\eg, GPT-4~\citep{GPT4}). Given each video segment and its associated textualized multimodal representations, we prompt the expert large language model to reason about why laughter occurred in the scene, considering diverse range of laughters such as hilarious, polite, or awkward laughter. Then the pseudo-label is generated, and these explanations serve as initial drafts of laughter reasoning annotations. 
% The prompt for generating pseudo-label through GPT-4 is given at Appendix~\ref{par:appendix_text_multimodal}.

\paragraph{Human verification and annotation}%
We conduct human verification to validate and refine the pseudo-labeled laughter reasoning and to annotate laughter types. We recruit annotators through Amazon Mechanical Turk (AMT) for both tasks.
For laughter type classification, annotators are provided with video segments and definitions of each laughter type. Then annotators instructed to label the laughter with confidence score, to ensure consistent labeling. The example of AMT survey is given at Appendix~\ref{sec:appendix_human_study_example}.
For laughter reasoning, annotators are instructed to watch the video segments, review the pseudo-generated explanations, and refine or correct them as needed. This verification step ensures both the accuracy and naturalness of the final annotations.

In total, our curated SMILE-Next comprises 3,590 video clips with 6,386 question–answer pairs, each annotated with multimodal textual representations, as summarized in \Tref{tab:smilenext_statistics}. The detailed statistics are provided at~Appendix \ref{sec:appendix_smilenext_statistics}. SMILE-Next is approximately four times larger than SMILE~\citep{smile24hyun} and covers a broader range of scenarios and tasks, enabling more comprehensive and general laughter understanding.

\section{Method}
\label{sec:method}
We aim to build a laughter expert LLM, a unified framework for identifying and understanding laughter in social interactions, building upon our curated SMILE-Next. To expand the LLM's capabilities for laughter understanding, we introduce two key components:
(1) Laughter-specific Self-Instruction, which automatically scales the dataset by generating synthetic question–answer pairs based on the human-annotated SMILE-Next data to cover more diverse scenarios; and
(2) Mixture-of-Laugh-Experts (MoLE), a parameter-efficient tuning framework that assigns lightweight LoRA experts to distinct laughter-related competencies while sharing a common backbone model.

\subsection{Laughter-specific Self-Instruction}
\label{sec:LSI}
While SMILE-Next covers diverse laughter-related tasks and scenarios with rich human annotations, we further expand its scale and scenario diversity to improve generalization to real-world settings. Motivated by \citet{wang2022self}, we propose a self-instruction framework that synthesizes instruction-response pairs tailored for laughter understanding.  This framework leverages an external LLM (We use GPT-4 API)  to generate both instructions and responses. This approach facilitates a scalable data acquisition pipeline, enabling rich knowledge distillation through supervised fine-tuning without additional manual annotation. Details of the prompts used for laughter-specific self-instruction, along with generated examples, are provided in the Appendix~\ref{sec:appendix_selfinst}.

\paragraph{Task generation}%
We begin by bootstrapping three seed tasks from SMILE-Next: laughter detection, type classification, and reasoning, and prompt an external LLM $M$ to generate new laughter-related tasks. We use GPT-4~\citep{GPT4} as $M$. The synthesized tasks cover a broader range of laughter-related competencies, such as predicting laughter behavior, reactions or intensity, and emotion classification, thereby extending the laughter understanding capability beyond the three original task categories.

\paragraph{Question and answer generation}%
Given the generated tasks, we further prompt the same LLM $M$ to synthesize question–answer pairs for each task, grounded in diverse real-world laughter scenarios and social relationships. These include, for example, workplace hierarchies (\eg, boss–employee) and informal relationships (\eg, friends). We observe that the self-instruction process effectively generates variations in laughter reasoning across different social contexts, enhancing the dataset’s generality. The resulting synthetic pairs expand coverage beyond humor-driven laughter to include socially motivated laughter, such as nervous, forced, sarcastic, or tension-mitigating responses.

\begin{figure}[t]
    \centering
    \includegraphics[width=0.5\textwidth]{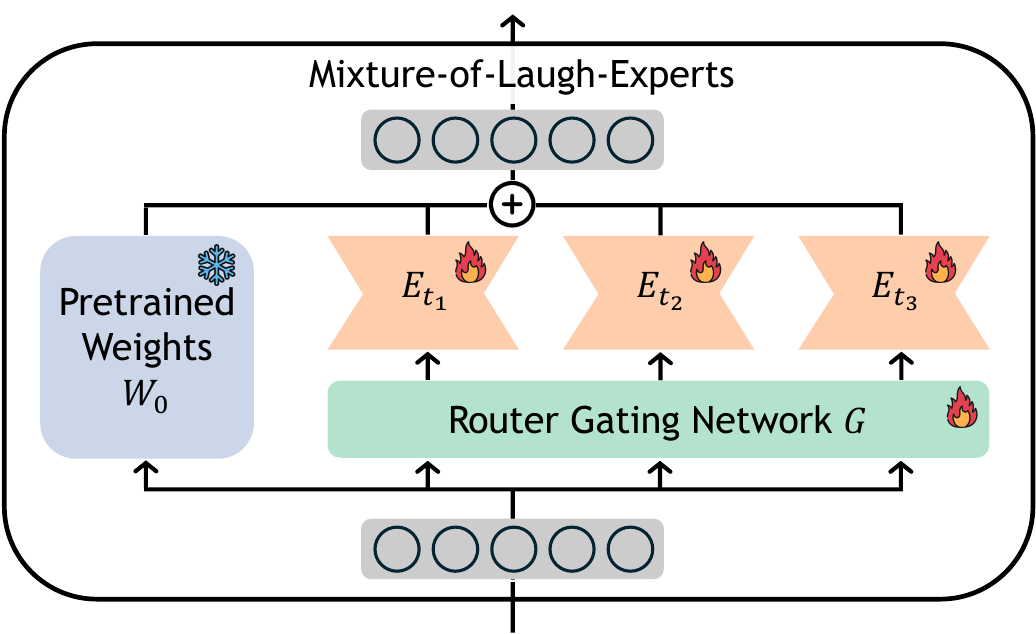}
    
    \caption{\textbf{Architecture of Mixture-of-Laugh-Experts.} 
    LoRA-based expert modules are added to a frozen pretrained weights and are dynamically weighted by a router gating network, resulting in laughter task-specific specialization.
    }
    % \vspace{-4mm}
    \label{fig:mole_architecture}
\end{figure}

\begin{table*}[t]  
  \centering
  \resizebox{1\linewidth}{!}{
    \begin{tabular}{@{}cccccccccc@{}}
    % \begin{tabular}{@{}ccp{2cm}p{2cm}p{2cm}p{2cm}cccc@{}}
    % \begin{tabular}{|c|c|c|c|c|c|c|c|c|c|}
    \toprule

    \multirow{2}{*}[-0.4em]{\textbf{Input Modality}} 
    & \multirow{2}{*}[-0.4em]{\textbf{Model}} 
    & \multicolumn{2}{c}{\textbf{Laughter Detection}} 
    & \multicolumn{2}{c}{\textbf{Laughter Type Classification}} 
    & \multicolumn{4}{c}{\textbf{Laughter Reasoning}} \\
    \cmidrule(r{2mm}l{2mm}){3-4} \cmidrule(r{2mm}l{2mm}){5-6} \cmidrule(r{2mm}l{2mm}){7-10}
    
    && F1 ($\uparrow$) & Acc. ($\uparrow$) 
    % & F1 ($\uparrow$) & Acc. ($\uparrow$)
    &\makebox[2cm]{F1 ($\uparrow$)} & \makebox[1cm]{Acc. ($\uparrow$)}
    & BLEU$_4$ ($\uparrow$) & METEOR ($\uparrow$) & ROUGE$_L$ ($\uparrow$) & SentBERT ($\uparrow$)\\
    \cmidrule{1-10}

    \multirow{2}{*}{\makecell{Audio-Visual LLM}} 
      & MiniCPM-o-v2.6     & 0.1952 & 0.3652 & 0.1919 & 0.2362 & 0.0802 & 0.1185 & 0.2632 & 0.5732 \\
      & Qwen2.5-Omni-7B    & 0.8373 & 0.8344 & 0.5856 & 0.5802 & 0.1683 & 0.1919 & 0.3549 & 0.7221 \\
    \midrule

    \multirow{2}{*}{\makecell{Visual LLM}}
      & Qwen2.5-VL         & 0.8586 & 0.8674 & 0.6949 & 0.7402 & 0.1665 & 0.1764 & 0.3284 & 0.6580 \\
      & Video-LLaVA        & 0.8692 & 0.8739 & \uline{0.7589} & 0.7912 & 0.1769 & 0.1818 & 0.3419 & 0.6567 \\
    \midrule

    \multirow{3}{*}{\makecell{LLM (Our approach)}}
      & Vicuna-v1.5        & \textbf{0.9675} & \textbf{0.9696} & 0.7460 & \uline{0.7953} & \uline{0.2354} & \uline{0.2311} & \textbf{0.4191} & \textbf{0.7907} \\
      & LLaMA3             & \uline{0.9674} & \textbf{0.9696} & \textbf{0.8067} & \textbf{0.8425} & \textbf{0.2427} & \textbf{0.2328} & \uline{0.4168} & \uline{0.7828} \\
      & Qwen2.5            & 0.9629 & \uline{0.9652} & 0.7094 & 0.7559 & 0.2303 & 0.2238 & 0.4097 & 0.7666 \\
    \bottomrule
    \end{tabular}
    }
    \caption{\textbf{Quantitative results.} 
    We report the result of SMILE-Next at diverse multimodal LLMs and LLMs. Utilizing our textualized multimodal cues, LLMs can outperform AV-LLMs and V-LLMs which utilizes implicit multimodal cues. 
    Models marked as Our Approach denote LLMs trained with the Laughter-specific Self-Instruction and Mixture-of-Laugh-Experts.
    }
  \label{tab:laugh_at_mix_task}
  \vspace{-2mm}
\end{table*}

\subsection{Mixture-of-Laugh-Experts}
\label{sec:MoLE}

Instead of training a single model to handle all laughter tasks, we propose the LoRA-based Mixture-of-Laugh-Experts (MoLE) framework which incorporates multiple trainable, lightweight expert modules. This design preserves the original LLM's reasoning capabilities while enabling task-adaptive expert specialization. Our approach explicitly decouples shared linguistic and reasoning knowledge from task-specific adaptations.

Figure~\ref{fig:mole_architecture} illustrates the MoLE framework. We freeze the pretrained weights $W_0$ of the backbone LLM to preserve its general reasoning knowledge and train only task-specific LoRA expert modules for each task, allowing the model to learn task-adapted behaviors efficiently. Such training strategy ensures versatility across a wide spectrum of task complexities, enabling the model to excel in diverse scenarios ranging from intricate reasoning to granular classification and detection.

For each laugher task $t\in \{t_{1}, t_{2}, t_{3}\}$, we initiate Laugh Expert $E_t$ to each task.
Each expert is injected to transformer's linear layer as an LoRA adapter.
Formally, for a linear transformation $W{\in}\mathbb{R}^{m\times n}$, $x \in \mathbb{R}^{n}$ represents the input (\ie, the previous hidden state) and $h$ as an output hidden state, LoRA reparameterizes it as
\begin{equation}
h=Wx = (W_0+BA)x,
\end{equation}
where $B\in\mathbb{R}^{m\times r}$ and $A\in\mathbb{R}^{r\times n}$.

To incorporate a task-specific expert for laughter understanding, we extend this formulation by introducing a mixture of LoRA experts. The residual adaptation term is replaced by a weighted combination of expert outputs:
\begin{equation}
h = W_0 x + \sum\nolimits_{i=1}^{T=3} R_i\cdot  E_i,
\end{equation}
where $T$ is the number of task-specific experts, $E_i$ is the output of the LoRA-adapted expert module for task $i$, and $R_i \in \mathbb{R}$ is the router gating weight assigned to expert $i$. 
The router gating weights $R_i$ are computed by first producing a gating weight, then normalized using a softmax function:
\begin{equation}
R_i = \mathrm{softmax}(W_g x)_,\quad E_i=B_iA_ix
\end{equation}
where $W_g \in \mathbb{R}^{T \times n}$ denotes learnable parameters and $B_i \in \mathrm{R}^{m\times r}, A_i \in\mathrm{R}^{r\times n}$ as an LoRA adapter. The resulting gating weights satisfy $\sum_{i=1}^{T} R_i = 1$ and enable \emph{soft routing}, allowing each input instance to be adaptively processed by a weighted combination of all experts.

As a result, each Laugh Expert $E_t$ learns task-specific specializations while enabling smooth interpolation across LoRA experts, which facilitates effective knowledge sharing across tasks while preserving task-specific expertise for laughter understanding. Combined with the pretrained LLM’s reasoning capabilities, we build an expert laughter understanding model that balances generalization and specialization.

\section{Experiments}
\label{sec:experiments}
\begin{figure*}[t]
    \centering
    \includegraphics[width=1\textwidth]{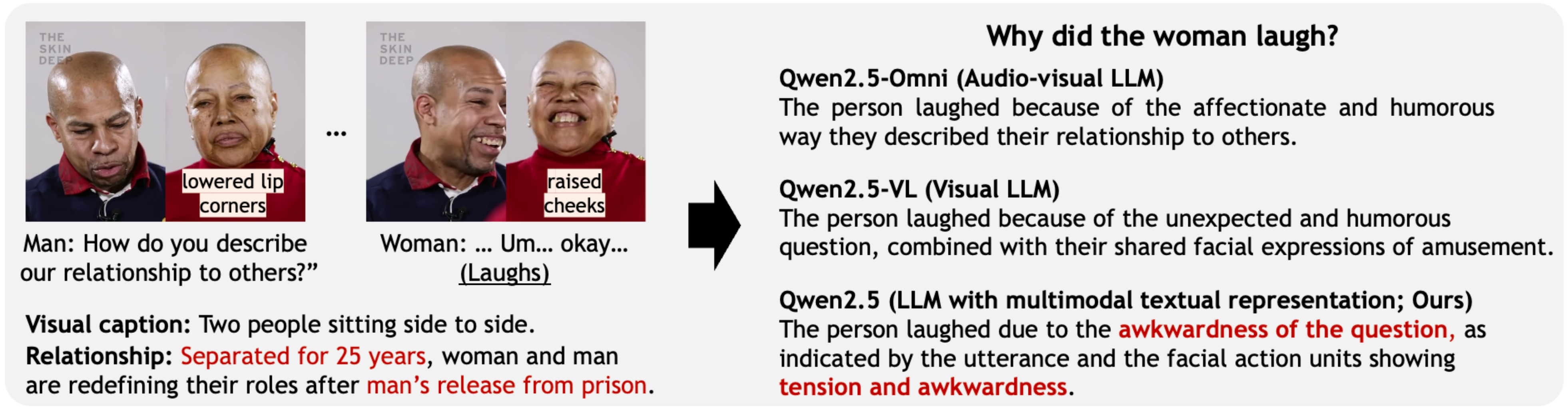}
    
    \caption{\textbf{Qualitative results.} As shown, LLMs trained at SMILE-Next can reason and analyze about the ambiguous and slight laughter, exactly giving the reason and classifying what this laugh is. Comparing AV-LLM and V-LLM to the LLMs which uses textual multimodal cues, it can explicitly point out to analyze why this laugh has occurred.
    }
    \label{fig:qual_result}
\end{figure*}

In this section, we describe the experimental setup and evaluate the performance of our proposed laughter expert LLM on the SMILE-Next dataset. We further present ablation studies to analyze the impact of our design choices.

\paragraph{Implementation details}%
Since our proposed self-instruction and Mixture-of-Laugh-Experts (MoLE) framework are model-agnostic, we evaluate our methods on three text-only LLMs: Vicuna-v1.5-7B~\citep{vicuna2023}, LLaMA3~\citep{grattafiori2024llama3}, and Qwen2.5~\citep{wang2024qwen2}. All models fully utilize the textualized multimodal inputs provided by the SMILE-Next dataset.
We integrate the MoLE into all linear layers of each LLM using LoRA adapters. We assign three experts, with the LoRA rank $r$ set to 8 and the scaling factor $\alpha$ set to 16 for each expert. The models are supervised fine-tuned with the SMILE-Next training set for three epochs with a batch size of 4, a learning rate of $1 \times 10^{-4}$, and a dropout rate of 0. Training is conducted using DeepSpeed ZeRO-3 for memory efficiency.

\paragraph{Baselines}%
We compare our approach against LLMs with varying multimodal configurations.
First, we evaluate audio-visual LLMs, specifically Qwen-Omni-7B~\citep{xu2025qwen2omni} and MiniCPM-o-v2.6~\citep{yao2024minicpm}, which process raw video and audio inputs directly without textualized cues. Second, we include visual LLMs such as Video-LLaVA~\citep{lin2023videollava} and Qwen2.5-VL~\citep{bai2025qwen2vl}; these models accept video inputs alongside textualized audio representations, including speech transcripts and relational descriptions.
All baselines are fine-tuned on the SMILE-Next for a single run training set using LoRA applied to all linear layers ($r=8, \alpha=16$) to ensure consistency.

\paragraph{Metrics}%
For laughter detection and type classification tasks, we report standard classification metrics, including precision, recall, F1-score, and accuracy.
For the laughter reasoning task, we evaluate generation quality using BLEU$_4$~\citep{papineni-etal-2002-bleu}, METEOR~\citep{banerjee-lavie-2005-meteor}, ROUGE$_L$~\citep{lin-2004-rouge}, and SentBERT~\citep{reimers2019sentence}.

\subsection{Experimental Results and Analyses}
\paragraph{Quantitative comparison}%
Table~\ref{tab:laugh_at_mix_task} summarizes the quantitative comparison across several types of LLMs that use different input representations. The results show that our proposed approach, leveraging LLMs with textualized multimodal representations achieves the best performance across all three tasks: laughter detection, classification, and reasoning, outperforming both audio-visual LLMs and visual LLMs. The text-only LLM backbones we evaluate show comparable performance trends, indicating that the gains primarily arise from the proposed representation and modeling strategy rather than from a specific backbone.
These findings suggest that our explicitly extracted and disentangled multimodal textual cues effectively support laughter understanding, consistent with observations in \citet{smile24hyun}. In contrast, models that directly process raw audio-visual or visual inputs must handle highly entangled signals, which makes it more challenging to capture the underlying causes of laughter.

\paragraph{Qualitative comparison}%
Figure~\ref{fig:qual_result} presents a qualitative example comparing laughter reasoning generated by an audio-visual (AV) LLM, a visual (V) LLM, and our proposed LLM. In this example, the woman laughs due to awkwardness and nervousness during the conversation. The AV-LLM interprets the scene as humorous, incorrectly conflating relational cues with spoken utterances. The V-LLM, which takes utterances and acoustic features as textual multimodal inputs, produces vague explanations and similarly misattributes the laughter to humor.
In contrast, our approach, an LLM leveraging all textualized multimodal cues captures the underlying awkwardness of the interaction and reflects it accurately in its reasoning by jointly considering relationship information and facial action units. These results demonstrate that extracting multimodal cues into textual representations enables more precise and robust laughter understanding in real-world settings.
Additional qualitative results are at Appendix~\ref{sec:appendix_additional_quals}.

{\begin{table}[t]
\centering
\resizebox{\columnwidth}{!}{
\begin{tabular}{lccc}
\toprule
\textbf{Pair (Left vs Right)} & \textbf{Left wins (\%)} & \textbf{Right wins (\%)} & \textbf{Tie} \\
\midrule
LLM (ours) vs V-LLM  & \textbf{55.7} & 37.4 & 7.1 \\
LLM (ours) vs AV-LLM & \textbf{69.0} & 26.2 & 4.8 \\
\midrule
 & \textbf{AV-LLM} & \textbf{V-LLM} & \textbf{LLM (ours)} \\
\midrule
Average Rank ($\downarrow$) & 2.38 & 1.93 & \textbf{1.69} \\
\bottomrule
\end{tabular}
}
\caption{\textbf{Human study of the textual multimodal cues.} As shown, humans prefer the answers of the proposed LLM, utilizing the textual multimodal cues instead of providing the entangled, raw signals.}
\label{tab:human_preference_evaluation}
\vspace{-4mm}
\end{table}

\paragraph{Human preference evaluation}%
\mok{We conducted a human preference evaluation of laughter reasoning task in which participants were shown responses from AV-LLM (MiniCPM-o), Vision-LLM (Qwen2.5-VL), and an LLM (LLaMA3; our setting), for each video sample and asked to rank them based on answer correctness given the video context.
As shown in Tab.~\ref{tab:human_preference_evaluation}, our LLM, which fully leverages textualized multimodal inputs, is consistently preferred over both V-LLM and AV-LLM, achieving a higher win rate (i.e., Left wins) and the lowest average rank. Notably, the strong preference for LLM suggests that textualized multimodal cues enable more effective laughter reasoning both better aligns with human judgments. Overall, these results demonstrate that analyzing and reasoning of laughter based on textual multimodal representations yields outputs that are more consistent with human interpretation.}

\begin{table}[t]  
  \centering
  \resizebox{1\linewidth}{!}{
    \begin{tabular}{ccccc}

    \toprule
    \multirow{2}{*}[-0.4em]{} & \multicolumn{4}{c}{\textbf{Laughter Detection}} \\
    \cmidrule(r{2mm}l{2mm}){2-5}
    
    & Precision ($\uparrow$)&Recall ($\uparrow$)&F1 ($\uparrow$) &Acc. ($\uparrow$) \\
    % \cmidrule{1-5}
    \midrule

    T & 0.8571 & 0.8669 & 0.8613 & 0.8696 \\
    T+A+V+R & \textbf{0.9582} & \textbf{0.9688} & \textbf{0.9629} & \textbf{0.9652} \\

    \midrule
    \multirow{2}{*}[-0.4em]{} & \multicolumn{4}{c}{\textbf{Laughter Type Classification}} \\
    \cmidrule(r{2mm}l{2mm}){2-5}
    & Precision ($\uparrow$)&Recall ($\uparrow$)&F1 ($\uparrow$) &Acc. ($\uparrow$) \\
    \midrule

    T & 0.6397 & 0.4436 & 0.4019 & 0.5433 \\
    T+A+V+R & \textbf{0.6828} & \textbf{0.7559} & \textbf{0.7094} & \textbf{0.7559} \\

    \midrule
    \multirow{2}{*}[-0.4em]{} & \multicolumn{4}{c}{\textbf{Laughter Reasoning}} \\
    \cmidrule(r{2mm}l{2mm}){2-5}
    & BLEU$_4$ ($\uparrow$)&METEOR ($\uparrow$)&ROUGE$_L$ ($\uparrow$) &SentBERT ($\uparrow$) \\
    \midrule

    T & 0.2047 & 0.2064 & 0.3862 & 0.7457 \\
    T+A+V+R & \textbf{0.2303} & \textbf{0.2238} & \textbf{0.4097} & \textbf{0.7666} \\
    \bottomrule
    \end{tabular}
    }
    \caption{\textbf{Ablation on multimodal information.} We conduct an ablation on each multimodal cues at Qwen2.5, and as shown utilizing all multimodal cues (transcription (T), acoustic cue (A), visual cue (V) and relation (R) enhances the performance of all laughter tasks: detection, type classification, and reasoning.}
  \label{tab:laugh_modality_ablation}
  \vspace{-3mm}
\end{table}

\begin{table}[t]
  \centering
  \resizebox{1\linewidth}{!}{
    \begin{tabular}{@{}l@{\,}c@{\,\,\,}c@{\,\,\,}c@{\,\,\,}c@{}}
    \toprule
    \multirow{2}{*}[-0.4em]{Model} & \multicolumn{4}{c}{\textbf{Laughter Detection}} \\
    \cmidrule(r{2mm}l{2mm}){2-5}
    
    &Precision ($\uparrow$)&Recall ($\uparrow$)&F1 ($\uparrow$) &Acc. ($\uparrow$) \\
    % \cmidrule{1-5}
    \midrule

    Vicuna-V1.5 & \underline{0.9724} & 0.9573 & 0.9645 & 0.9522 \\
    \ \textbf{(+Self-Instruction)} & \textbf{0.9912} & \underline{0.9709} & \textbf{0.9806} & \underline{0.9630} \\
    \ \textbf{(+MoLE)} & 0.9635 & \textbf{0.9722} & \underline{0.9675} & \textbf{0.9696} \\
    \midrule
    Qwen2.5 & 0.9550 & \underline{0.9684} & \underline{0.9608} & \underline{0.9636} \\
    \ \textbf{(+Self-Instruction)}  & \textbf{0.9599} & 0.9587 & 0.9589 & 0.9587 \\
    \ \textbf{(+MoLE)}  & \underline{0.9582} & \textbf{0.9688} & \textbf{0.9629} & \textbf{0.9652} \\

    \midrule
    \multirow{2}{*}[-0.4em]{Model} & \multicolumn{4}{c}{\textbf{Laughter Type Classification}} \\
    \cmidrule(r{2mm}l{2mm}){2-5}
    &Precision ($\uparrow$)&Recall ($\uparrow$)&F1 ($\uparrow$) &Acc. ($\uparrow$) \\
    \midrule

    Vicuna-V1.5 & 0.5164 & 0.5370 & 0.4892 & 0.6923 \\
    \ \textbf{(+Self-Instruction)} & \underline{0.7029} & \underline{0.7250} & \underline{0.7050} & \underline{0.7229} \\
    \ \textbf{(+MoLE)} & \textbf{0.7412} & \textbf{0.7608} & \textbf{0.7460} & \textbf{0.7953} \\
    \midrule
    
    Qwen2.5  & 0.6721 & \underline{0.7480} & \underline{0.7023} & 0.7480 \\
    \ \textbf{(+Self-Instruction)}& \textbf{0.7066} & 0.6809 & 0.6852 & \textbf{0.7953} \\
    \ \textbf{(+MoLE)}& \underline{0.6828} & \textbf{0.7559} & \textbf{0.7094} & \underline{0.7559} \\

    \midrule
    \multirow{2}{*}[-0.4em]{Model} & \multicolumn{4}{c}{\textbf{Laughter Reasoning}} \\
    \cmidrule(r{2mm}l{2mm}){2-5}
    &BLEU$_4$ ($\uparrow$)&METEOR ($\uparrow$)&ROUGE$_L$ ($\uparrow$) &SentBERT ($\uparrow$) \\
    \midrule

    Vicuna-V1.5 & 0.1872 & 0.1869 & 0.3424 & 0.6999 \\
    \ \textbf{(+Self-Instruction)} & \underline{0.2073} & \underline{0.2110} & \underline{0.3840} & \underline{0.7423} \\
    \ \textbf{(+MoLE)} & \textbf{0.2354} & \textbf{0.2311} & \textbf{0.4191} & \textbf{0.7907} \\
    \midrule
    
    Qwen2.5  & 0.2272 & 0.2227 & \underline{0.4053} & 0.7640 \\
    \ \textbf{(+Self-Instruction)} & \underline{0.2289} & \underline{0.2250} & 0.4045 & \textbf{0.7704} \\
    \ \textbf{(+MoLE)} & \textbf{0.2303} & \textbf{0.2238} & \textbf{0.4097} & \underline{0.7666} \\
    \bottomrule
    \end{tabular}
    }
    \caption{\textbf{Ablation on Laughter-specific Self-Instruction \& MoLE.} Sequentially incorporating Self-instruction and MoLE improves overall performance across the three laughter understanding tasks.}
    \vspace{-2mm}
  \label{tab:selfinst_ablation}
\end{table}

\subsection{Ablation Studies}

\paragraph{Multimodal cues}%
Table~\ref{tab:laugh_modality_ablation} analyzes the impact of different textual multimodal cues. We compare models using only text transcripts (T) against those incorporating the full set of information: transcripts (T), acoustic (A), visual (V), and relational (R) cues. The results show that relying solely on textual utterances is insufficient for robust laughter understanding, as laughter can arise from a wide range of non-verbal signals, such as facial expressions, vocal pitch and tone, background context, and interpersonal relationships. This demonstrates that jointly considering contextual, acoustic, and visual cues, and relational cues is essential for robust laughter understanding.

\paragraph{Effect of self-instruction and MoLE}%
As shown in \Tref{tab:selfinst_ablation}, self-instruction consistently improves performance across all three tasks for most LLM backbones. In particular, for the laughter reasoning task, self-instruction helps preserve the model's reasoning capabilities while simultaneously improving detection and classification performance. This suggests that self-instruction effectively bridges gaps in real-world scenario diversity that cannot be fully captured by human-annotated data alone. Furthermore, incorporating the MoLE framework yields further performance gains across tasks. This improvement indicates that introducing multiple lightweight experts enables the model to better specialize in distinct laughter-related competencies while sharing a common backbone.

\paragraph{Analysis on MoLE router weights}%
Figure~\ref{fig:router_weight} visualizes the MoLE router weights and reveals task-dependent expert activation patterns across laughter understanding tasks. While Expert~1 remains dominant across all three tasks, the relative contributions of the other experts vary depending on the task. In particular, laughter type classification assigns more weight to Expert~2, while laughter detection assigns slightly more weight to Expert~3, whereas laughter reasoning relies more heavily on Expert~1. These variations indicate that the MoLE framework dynamically allocates expert capacity according to task-specific requirements, rather than uniformly activating a single expert.

\paragraph{Noise robustness in multimodal inputs}%
\mok{We evaluated the model’s robustness to noise in textualized multimodal inputs by perturbing a single modality (facial action units or acoustic features) while keeping others unchanged.
The model maintains stable reasoning under moderate noise, up to~30\% perturbation in facial features and ~40–50\% noise in acoustic signals. Beyond these thresholds, performance degrades, leading to incorrect or unreliable interpretations (e.g., misjudging pitch).
We assume that such robustness happened to our model due to the ability of leveraging multimodal complementarity, down-weighting corrupted inputs and relying on remaining modalities to sustain accurate reasoning.}

\begin{figure}[t]
    \centering
    \includegraphics[width=0.48\textwidth]{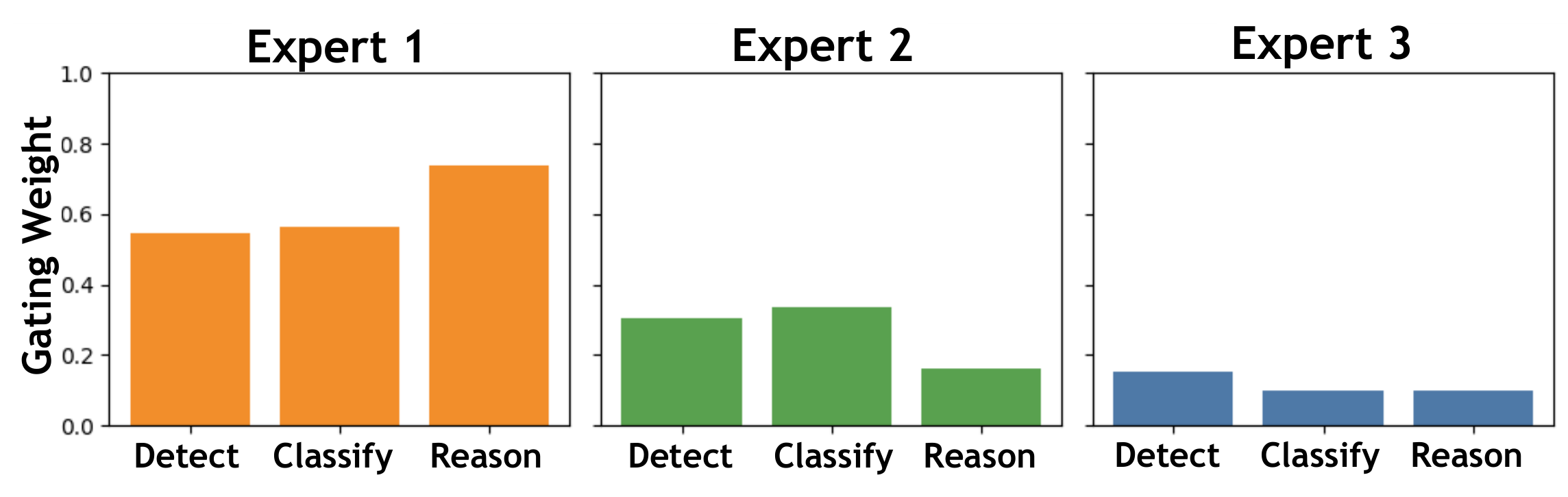}
    \caption{\textbf{Visualization of MoLE's router weights.} The results show dominant activation of Expert~1, with task-dependent variations, classification assigns more weight to Expert~2, while detection does so for Expert~3.}

    \label{fig:router_weight}
    \vspace{-3mm}
\end{figure}

\paragraph{Computational Latency of MoLE}%
\mok{We report the latency of proposed architecture MoLE at Table~\ref{tab:latency}.
MoLE introduces only a marginal latency increase of approximately 0.02 seconds per sample compared to the single-expert model, corresponding to about 1–2\% additional inference time, which is negligible in practice. Rather than latency reduction, MoLE enhances laughter understanding while incurring minimal computational overhead.}

\begin{table}[t]
\centering
\resizebox{\columnwidth}{!}{
\begin{tabular}{lcccc}
\toprule
 & \textbf{Detect. (ms)} & \textbf{Cls. (ms)} & \textbf{Reasoning (ms)} & \textbf{All (ms)} \\
\midrule
Single expert & 981 & 790 & 2802 & 1494 \\
Multi-experts(MoLE) & 991 & 796 & 2845 & 1513 \\
Difference & \textbf{+10} & \textbf{+6} & \textbf{+43} & \textbf{+19} \\
\bottomrule
\end{tabular}
}
\caption{\textbf{Latency comparison between single expert and multi-expert (MoLE)}. We report the latency of single-expert and proposed multi-expert (MoLE). MoLE demands only marginal latency, about 1-2\% additional time.}
\label{tab:latency}
\vspace{-3mm}
\end{table}

\section{Conclusion}
\label{sec:conclusion}
In this work, we introduce SMILE-Next, an enhanced multimodal dataset for real-world laughter understanding that covers three laughter-related taskss: laughter detection, laughter type classification, and laughter reasoning. 
Based on the SMILE-Next, we propose a unified framework for laugh expert LLM that combines laughter-specific self-instruction with a Mixture-of-Laugh-Experts modeling approach to enable effective laughter understanding in LLMs. 
By textualizing multimodal signals, our framework captures critical cues underlying laughter, consistently outperforming raw multimodal representations in extensive experiments. 
Moreover, by jointly modeling multiple laughter-related tasks with parameter-efficient fine-tuning, we develop a robust laughter expert LLM for real-world scenarios. 
Overall, this work advances the understanding of complex human behaviors with a particular focus on laughter in natural social interactions.

\section*{Limitation}%
SMILE-Next sources videos from a wide range of scenarios, still it may not fully capture all possible forms of laughter occurring in real-world settings. Certain rare or atypical laughter cases, such as culturally specific expressions, nuanced social norms, or complex group interactions may be underrepresented. In addition, the dataset primarily focuses on English-language content, reflecting the dominance of English-speaking sources, and therefore does not adequately cover laughter expressions in other languages or cultural contexts. Future work will explore incorporating multilingual and cross-cultural laughter data, along with more complex social interactions, to further improve coverage and generalization.

\section*{Ethical Considerations}
SMILE-Next is curated exclusively from publicly available videos, with data usage governed by the original platforms’ policies.
While laughter understanding can enable more natural human–AI interactions, misinterpretation or misuse of such system can raise concerns related to bias or unintended social consequences. We encourage responsible use of this dataset and associated models, with careful consideration of ethical implications. 
AI assistants were used only for grammatical correction and writing refinement.

\newpage
\section*{Acknowledgment}
This work was partially supported by InnoCORE program  (N10250156, KAIST InnoCore LLM (25\%); % 이노코어 LLM
26-InnoCORE-01, Trust-Enhanced Mutualistic Bio-Embedded AI (25\%)), % 바이오체화 LLM
Institute of Information \& communications Technology Planning \& Evaluation (IITP) (No. RS-2024-00457882, National AI Research Lab Project) (25\%), % AI 거점
and the National Research Foundation of Korea (NRF)  (No. RS-2024-00451947) (25\%) grants funded by the Korea government (MSIT). % NRF
This work is also funded by Toonation Co. % 투네이션

% \newpage
\bibliography{ref}
% \nocite{apte1985humor}

\clearpage
\newpage
\appendix
\section{Appendix}
\label{sec:appendix}
\subsection{Additional Explanation of SMILE-Next}\label{sec:appendix_smilenext_statistics}
In this section, we provide additional statistics and examples of SMILE-Next. First, we provide how our SMILE-Next is constructed, second, we provide how each laughter task is prompted to our laughter-expert LLM.

\paragraph{Dataset statistics}%
SMILE-Next contains three tasks: laughter detection, type classification, and reasoning. \Tref{tab:smilenext_statistics} reports the train/val/test splits per each task. Furthermore, 
we provide the statistics of (1) Number of utterances (2) Video duration (3) Length of Instruction (4) Length of Answer at~\Fref{fig:appendix_smilenext_statistics}. As shown, SMILE-Next covers 2-8 utterances at most, and for video duration, it covers around 30-50 seconds. For length of instruction, since we feed the multimodal cues in textualized form, most samples require 2000-4000 letters, and output answers are mostly around 150-250 letters.

\paragraph{Constructing QA pairs}
Since we formulate each task into question-answering format, we provide the prompts for constructing the instruction dataset for each tasks below.

For laughter detection, we constructed the question instruction as: \textit{"Detection task: You are a laugh detector. Find out if there is laugh in this clip."}
And we forced model to respond either \textit{"Yes, there is laugh in this video."} or \textit{"No, there is no laugh in this video."}. 
For laughter type classification, we constructed the question instruction as: 
\textit{"Type Classification task: you are to answer the class of laugh type. There are three types: Polite, Satirical, Mirthful. \{description of each type\}"}, and we forced the answer to be \textit{"The laugh type is \{laughter type\}"}.
For laughter reasoning, we constructed the instruction as: \textit{"Reasoning task: you are to answer why the person laughed at most 30 words, starting with "The audience/person laughed because'".}
So the output format is fixed as \textit{"The audience/person laughed because"}, and conducting the reasoning.

\subsection{Detailed Dataset Construction Pipeline}\label{sec:appendix:dataset_construction}
In here, we provide the details of SMILE-Next's dataset construction pipeline, at each stage (1) Laughter video collection (2) Textual multimodal information extraction (3) Pseudo-label annotation (4) Human verification and annotation.

\paragraph{Laughter video collection}
We collected the video clips from YouTube 'The Skin Deep' and 'MovieClips'. The duration of each sources' video clip is different, because the type of conversation from the source is different. 
Similar to SMILE~\citep{smile24hyun}, we targeted the last laughter happening in the video clip, since the video clip may contain multiple laughs. Also focusing on the `laugh happening scene', we did not annotate who laughed. We assumed the laughter happens contagiously in dyadic conversation which makes the other person also laugh, making the whole atmosphere to laughing scene.

\begin{figure}[t]
    \centering
    \includegraphics[width=0.5\textwidth]{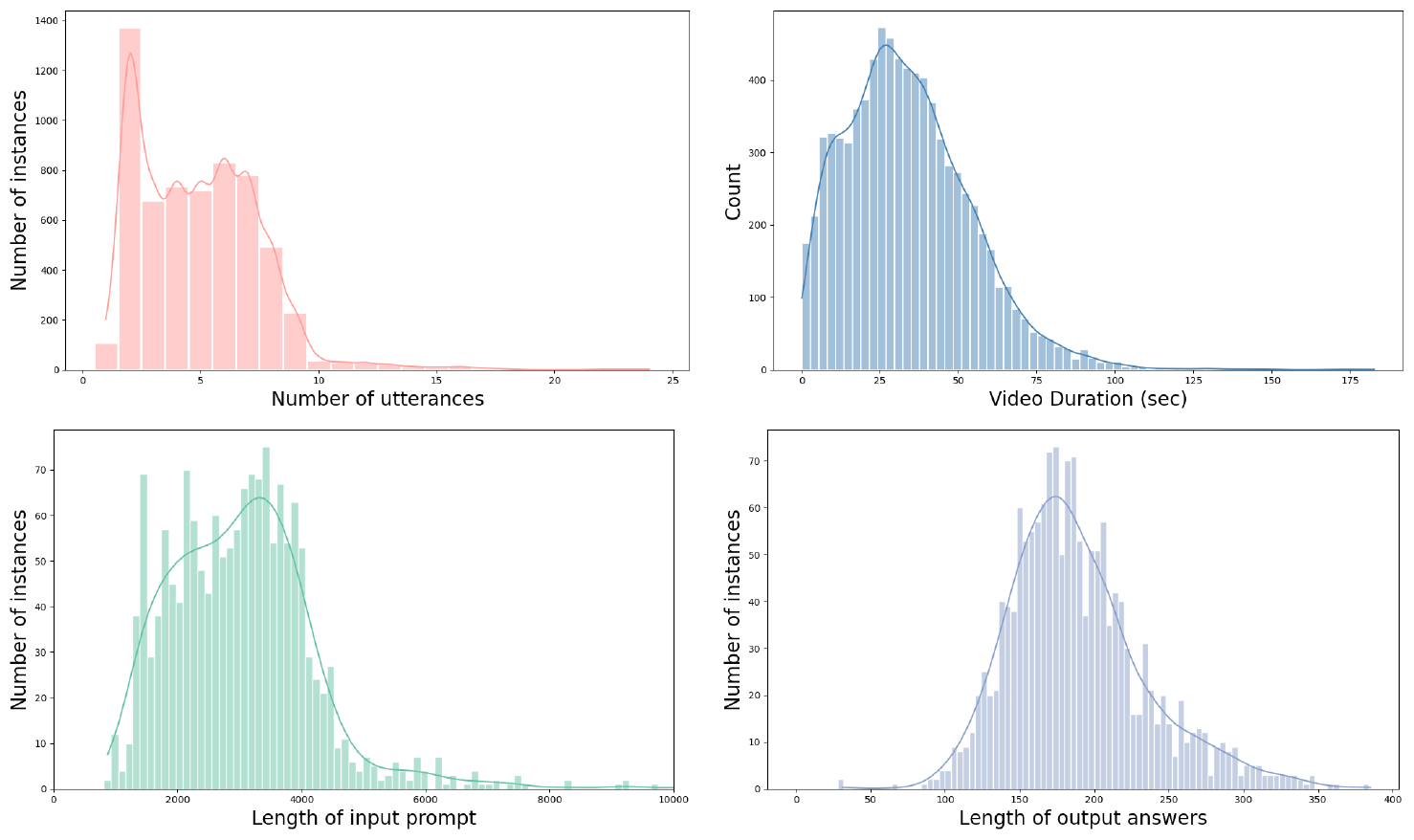}
    
    \caption{\textbf{SMILE-Next statistics.} We report the number of utterances and the video duration of each video samples in SMILE-Next at top row, which shows similar distribution. At bottom row, we also report the length of input and output text, and as shown input prompt is diverse while the output text shows distribution centered at 200.
    }
    \label{fig:appendix_smilenext_statistics}
\end{figure}

\begin{table*}[t]
  \centering
  \begin{tabular}{p{0.98\textwidth}}
    \toprule
    \textbf{Prompt for GPT-4 Generating Pseudo-Reasoning Labels} \\
    \midrule
    Reasoning task: you are to answer why the person laughed given the video clip. The video clip from
    the youtube channel 'The Skin Deep', titled {title}, with multimodal information (Utterance, Facial Action Units, Video caption, Relation, Clip description,
    Acoustic features (10 dimension; 1.mean of F0 contour, 2.var of F0 contour, 3. mean of energy contour, 4. mean of voiced duration, 5. mean of unvoiced duration, 6. var of 
    energy contour, 7. first derivative of fundamental frequency, 8. second derivative of fundamental frequency, 9. jitter, 10. shimmer)) is given.
    Referring to the multimodal information (Utterance, Prosodic, Facial Action Units, Video caption, Topic) in each utterance, 
    Please make sure that there are a lot of laughters, such as hilarious laugh polite laugh, awkward laugh, dumbfound laugh, etc.
    Try to detect and reason possible laughter. Based on the given query, give reason about why the person laughed, at most 30 words, starting with 'the person laughed'.
    For this, please state which part of multimodal information (utterance, Prosodic, Facial Action Units, Video caption, Topic, relation) has stimulated the laugh.
    For example, utterance "I am an idiot" can be a first starting point for laughter reasoning. 
    If there is not enough context, try to focus on relation ot clip description, those can also be a starting point to reason for laugh.
    Given video clip: \{Textualized multimodal information\}\\
    \bottomrule
  \end{tabular}
  \caption{\textbf{Prompt for pseudo-reasoning label annotation.} We provide the prompt for generating the pseudo-reasoning label. The prompt is fed to GPT-4 to generate free-form laughter reasoning.}
  \label{tab:pseudo_reasoning_label_prompt}
\end{table*}

\paragraph{Textual multimodal information extraction}\label{par:appendix_text_multimodal}
We report the off-the-shelf models for extracting multimodal information.
For the utterance cue acquisition, we have leveraged WhisperX~\citep{bain2023whisperx} and trimmed a video into multiple segments; one utterance for one segment. For each segments, we also leveraged Active Speaker Algorithm~\citep{tao2021someone} using audio-visual information in each video.
For the visual cue, we acquired video caption for each segment, which explains each segment's situation, providing the visual cue. We also extracted the facial action units~\citep{yao2021action} of speakers in the video; Accumulating each frame's facial action and selected most dominant top-3 facial action units for each person.
We extract acoustic features from speech utterances using off-the-shelf speech processing tools~\citep{dehak2007modeling}, including the mean and variance of pitch and intensity, as well as jitter and shimmer.
Although we initially explored transforming these numerical features into categorical linguistic descriptors (e.g., labeling a mean pitch above 200 Hz as “high pitch”), we found this approach to be problematic due to the inherently subjective nature of such mappings.
So we retain the raw numerical values and directly incorporate them as acoustic features. These statistics are conveyed to the LLM via natural-language prompts that describe the feature values, enabling the laugh LLM to interpret them using its prior knowledge of speech and prosodic patterns.

\paragraph{Pseudo-label annotation}
For generating pseudo-reasoning labels, we prompted an expert large language model (LLM) to generate pseudo-laughter reasoning. 
We used GPT-4 as an expert LLM, and the prompt is given at ~\Tref{tab:pseudo_reasoning_label_prompt}. As shown, we provide the detailed description about given textual multimodal cues, including utterance, facial action unit, video caption, relation and clip description. 
We further prompt the model to generate reasoning that explicitly references the multimodal cues responsible for inducing laughter, encouraging grounded and interpretable explanations. This design allows the model to produce coherent and context-aware pseudo-reasoning annotations aligned with real-world social interactions.
Annotations that passed human verification were retained as-is, while those that did not were further refined by human annotators and subsequently used in the dataset.

\paragraph{Human verification and annotation} \label{sec:appendix_laughter_type_reason_annotation}
For the annotation of the laughter, we ran the survey on Amazon Mechanical Turk (AMT) for labeling the reason and type of laughter corresponding to its confidence. 
We have given the precise instruction to start reasoning, for example, when some laughter just happens, without an explicit reason, we instructed annotators to make the reason of laughter considering the laughter as the `reaction' of some explicit cue (\eg utterance, relation)
For laughter type classification, annotators were asked to watch a short video segment containing laughter and assign one of the five labels based on the contextual and acoustic characteristics of the laughter. To ensure annotation quality, 3 annotators were assigned per instance. 
After the annotation, we select the most-voted laughter type as the GT, considering the confidence score of the selected laughter type. 
We add the screenshot of laughter type classification AMT survey at~\Sref{sec:appendix_human_study_example}.
\mok{We measured inter-annotator agreement using Fleiss’ Kappa (0.42195). Despite the task’s inherent subjectivity, this score indicates moderate agreement beyond chance, suggesting that the labeling guidelines ensured consistency and supporting the reliability of the annotations.}

\begin{table}[t]  
  \centering
  \resizebox{1\linewidth}{!}{
    \begin{tabular}{lcccc}
    \toprule
    \multirow{2}{*}[-0.4em]{Model} & \multicolumn{4}{c}{\textbf{Laugh Detection}} \\
    \cmidrule(r{2mm}l{2mm}){2-5}
    
    &Precision ($\uparrow$)&Recall ($\uparrow$)&F1 ($\uparrow$) &Acc. ($\uparrow$) \\
    % \cmidrule{1-5}
    \midrule

    Video-LLaVA (Vicuna-V1.5) & 0.859 & 0.882 & 0.865 & 0.870 \\
    Vicuna-V1.5 & \textbf{0.922} & \textbf{0.915} & \textbf{0.913} & \textbf{0.915} \\

    \midrule
    \multirow{2}{*}[-0.4em]{Model} & \multicolumn{4}{c}{\textbf{Laugh Type Classification}} \\
    \cmidrule(r{2mm}l{2mm}){2-5}
    &Precision ($\uparrow$)&Recall ($\uparrow$)&F1 ($\uparrow$) &Acc. ($\uparrow$) \\
    \midrule

    Video-LLaVA (Vicuna-V1.5) & 0.713 & 0.519 & 0.589 & 0.529 \\
    Vicuna-V1.5 & \textbf{0.725} & \textbf{0.692} & \textbf{0.660} & \textbf{0.692}\\

    \midrule
    \multirow{2}{*}[-0.4em]{Model} & \multicolumn{4}{c}{\textbf{Laugh Reasoning}} \\
    \cmidrule(r{2mm}l{2mm}){2-5}
    &BLEU$_4$ ($\uparrow$)&METEOR ($\uparrow$)&ROUGE$_L$ ($\uparrow$) &SentBERT ($\uparrow$) \\
    \midrule

    Video-LLaVA (Vicuna-V1.5) & 0.176 & 0.181 & 0.342 & 0.656 \\
    Vicuna-V1.5 & \textbf{0.210} & \textbf{0.213} & \textbf{0.379} & \textbf{0.745} \\
    \bottomrule
    \end{tabular}   
    }
    \caption{\textbf{Quantitative results of implicit and textual cues with sample backbone.} We evaluate encoder-based and textualized multimodal representations under an identical Vicuna-7B-v1.5 backbone for laughter detection, type classification, and reasoning, and textualized multimodal cues outperform encoder-based multimodal cues.}
    % \vspace{-1mm}
  \label{tab:laugh_at_each_task}
\end{table}

\subsection{Additional Quantitative Result: Comparison of Implicit and Textual Cues}\label{sec:laugh_at_each_tasks}

\paragraph{Result}
To verify the effectiveness of our textualized multimodal representation, we compare two models that share the same backbone (Vicuna-7B-v1.5): (1) a multimodal LLM that encodes multimodal inputs using feature encoders, and 
(2) an LLM that consumes the proposed textualized multimodal information. In here, we evaluate using the captioning metrics: $\text{BLUE}_4$, METEOR, $\text{ROUGE}_L$ and SentBERT for all three tasks.
We report the result of each tasks at~\Tref{tab:laugh_at_each_task}.
Across all three laughter-related tasks: laughter detection, laughter type classification, and laughter reasoning, the text-based model consistently outperforms its encoder-based counterpart.
Notably, the performance gains are observed not only in low-level perception tasks such as detection and type  classification, but also in the higher-level reasoning task, indicating that textualized multimodal information enables the language model to more effectively integrate and reason over multimodal cues.
These results suggest that explicitly textualizing multimodal signals provides a more accessible and coherent interface for LLMs than relying on modality-specific encoders, particularly for complex social reasoning tasks such as laughter understanding.

\paragraph{Understanding Performance Gains in Laughter Detection}
\begin{table}[t]
\centering
\resizebox{\linewidth}{!}{
\begin{tabular}{l c}
\toprule
\textbf{Model} & \textbf{Acc (\%) at UR-Funny} \\
\midrule
TFN (ACL '17) & 64.7 \\
MISA (ACM MM '20) & 70.6 \\
SMILE (NAACL '24) & 79.0 \\
\textbf{SMILE-Next (Ours)} & \textbf{92.8} \\
\bottomrule
\end{tabular}
}
\caption{\textbf{Performance comparison on the UR-Funny dataset.} As shown, the accuracy of laughter detection on UR-Funny has steadily improved over time, with recent models achieving strong performance, indicating the increasing maturity of the task.}
\label{tab:urfunny_results}
\end{table}
\mok{While laughter detection performance has recently approached saturation, as shown at Tab.~\ref{tab:urfunny_results}, this trend reflects model maturity rather than a limitation of the task or dataset. As model capabilities have advanced, performance on benchmarks such as UR-Funny~\cite{hasan-etal-2019-ur} has steadily improved, with our model achieving 92.8\% accuracy, substantially outperforming prior approaches. We interpret this result as evidence that laughter detection, as currently formulated, is becoming a well-solved problem. Consequently, this motivates the need to move beyond binary detection toward more nuanced and cognitively meaningful tasks. To this end, we introduce additional tasks, laughter type classification and laughter reasoning to capture richer aspects of laughter understanding and to encourage future research in this direction.}

\paragraph{Error analysis}%
\mok{We additionally qualitatively identify recurring failure patterns in the model’s predictions. First, the model occasionally confuses mirthful and polite laughter, particularly in cases where genuine amusement emerges from shared memories or close interpersonal contexts, where expressive cues tend to be subtle and ambiguous. Second, the model exhibits occasional over-reliance on multimodal signals, over-interpreting facial or acoustic cues even when the textual utterance alone provides sufficient evidence for accurate reasoning.}

\subsection{Additional Result for Laughter-specific Self-Instruction}\label{sec:appendix_selfinst}

\begin{figure}[t]
    \centering
    \includegraphics[width=0.50\textwidth]{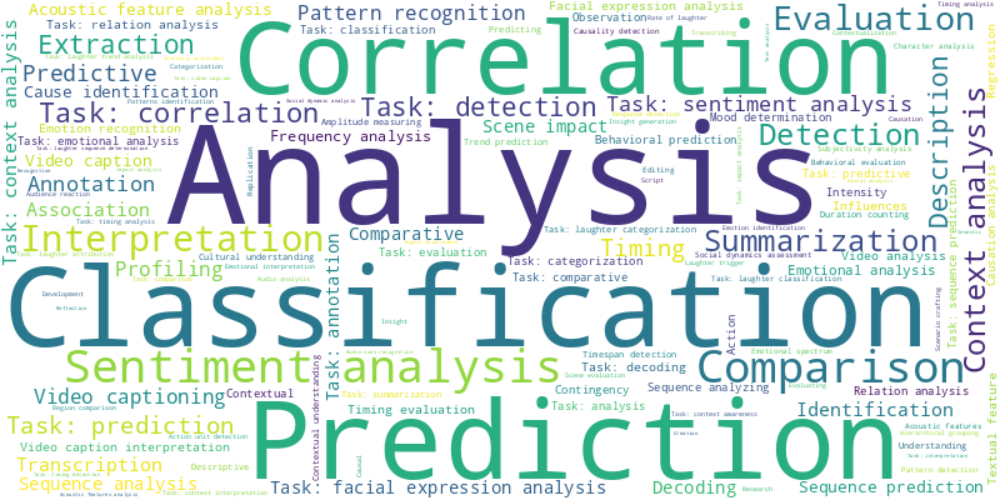}
    \caption{\textbf{Wordcloud for generated self-instruction tasks.} As shown, the generated tasks frequently contains the analysis, classification and prediction tasks, such as sentiment analysis, or context analysis, and laughter-situation predictions.}
    \label{fig:selfinst_wordcloud}
\end{figure}

\begin{table}[t]
\centering
\resizebox{0.95\linewidth}{!}{
\begin{tabular}{ccc}
\hline
\toprule
\textbf{top-k}         & \textbf{Task name} & \textbf{Counts} \\
\midrule
1             & Analysis Task  &    31  \\ 
2             & Classification Task  &    30  \\ 
3             & Prediction Task   &    29  \\ 
4             & Correlation Task  &    24  \\ 
5             & Sentiment Analysis Task & 22 \\ 
% 6             & Comparison Task     & 21 \\ 
% 7             & Interpretation Task & 20 \\ 
% 8             & Evaluation Task & 17 \\ 
% 9             & Summarization Task & 13 \\
% 10            & Context Analysis Task & 13 \\ 
\bottomrule
\end{tabular}
}
\caption{\textbf{Top-5 generated tasks through laugh self-instruction.} We have parsed each generated instances' task and counted which tasks are generated. As shown, newly generated tasks contains various of task, containing correlation or sentiment analysis tasks.}
\label{tab:selfinst_topk}
% \vspace{-4mm}
\end{table}

Through laughter-specific self-instruction, we generate 1,790 samples containing diverse laughter-related tasks at real world scenarios. Those samples are added as a dataset with the curated SMILE-Next dataset during supervised fine-tuning, ensuring the model to deal with diverse laughter-inducing senarios with diverse laughter related tasks,
leading to a more robust and generalized laughter understanding LLM.

\paragraph{Analysis of generated tasks}
We analyze what tasks are generated from the Laughter-specific Self-Instruction. First, we report the wordcloud of the task names generated from self-instruction at~\Fref{fig:selfinst_wordcloud}. As shown, it contains diverse tasks including analysis, classification, prediction, and correlation. Based on SMILE-Next's three core tasks, such a self-instruction method diversifies the number of tasks with generic laughter situations. 
Second, we report the top-5 tasks generated through the self-instruction at Table~\ref{tab:selfinst_topk}. We have clustered the self-instruction-generated tasks by parsing and renaming, and, as shown, this has yielded around 300 new laughter-related tasks, increasing diversity. So, enhancing our manual laughter-related task, self-instruction of the laughter dataset guarantees the diversity.

\paragraph{Implementation of Laughter-specific Self-Instruction}%
Our implementation follows the official Self-Instruction framework~\citep{wang2022self}. Building upon this foundation, we extend the original setup by incorporating laughter-specific prompts to generate new tasks and corresponding instances tailored to laughter understanding.
These prompts are designed to elicit diverse laughter-related behaviors and reasoning patterns, thereby complementing the supervision provided by SMILE-Next.
The prompts used for task and instance generation are shown in \Tref{tab:selfinst_prompt_and_examples}. We explicitly encouraged the model to generate not only amusement-driven laughter but also socially motivated ones such as forced, nervous, or sarcastic laughter, ensuring that the laughter-specific Self-Instruct framework captures a diverse range of laughter-inducing scenarios.

\paragraph{Examples of Laughter-specific Self-Instruction}
We report the examples of generated self-instruction samples at ~\Tref{tab:selfinst_prompt_and_examples}. 
As shown at \Tref{tab:selfinst_prompt_and_examples} below of changing the laughter reasoning, according to its relations, for example when the situation is that boss is giving the funny joke, and the employees laughed, and our synthesized data pinpoint their relations to reason for the laugh, saying that it is a forced laugh that laugh happened due to its forced situation, not because the joke is funny. This enhances our laughter dataset to be more generalized to the real situation, while our annotated dataset considers only the partial part of the real-world situations.

\begin{table*}[t]
  \caption{\textbf{Prompt for Laughter-specific Self-Instruction and Examples of Generated Self-Instruction Instances.} 
  } 
  \centering
  \resizebox{1\linewidth}{!}{
    \begin{tabular}{p{0.95\linewidth}} % 너비 지정
    \hline
    \toprule
    \textbf{Laughter-specific Self-Instruction Prompts} \\
    \midrule
    \textbf{Task Generation: }Come up with a series of tasks:\\
    \{few shot examples\} \\
    \midrule
    \textbf{Instance Generation}: Generate new examples that follow the same format as above. Include a variety of laughter situations, not only humorous or joyful ones, but also socially-driven laughs such as forced, nervous, or sarcastic laughter. Ensure the context reflects subtle multimodal cues such as facial expression, tone, or social dynamics.\\
    \{few shot examples\} \\
    \hline
    \toprule
    \textbf{Examples of Generated Self-Instruction Instances} \\
    \midrule
    \textbf{Evaluating task:} Rate each laugh in the scene based on intensity and context, and determine whether it was genuine or forced.\\[1mm]
    \textbf{Input:} During a tense board meeting at her office, Sarah tries to lighten the mood with a faint chuckle after the boss makes a dry joke. Her co-workers don't seem to respond much. \\
    \textbf{Answer:} Forced, low intensity \\
    \midrule
    \textbf{Correlation task: } Derive the relationship between the acoustic features and the intensity or type of laugh.\\
    \textbf{Input:} Acoustic feature: irregular pace, variation in pitch (Laughter)\\
    \textbf{Answer:} This could suggest a nervous laughter or possibly a fake laugh.\\
    \midrule
    \textbf{Laugh pattern analysis task: } Analyze the recurring patterns, if any, in the laughter based on the context, relationship, and emotional state of the people involved. \\
    \textbf{Input:} Shawn and Brooke are best friends. They often tend to laugh when shared jokes or funny moments between them, creating a contagious laughter pattern that generally initiates with Shawn's giggle followed by Brooke's loud laughter.\\
    \textbf{Answer:} The recurring pattern observed in Shawn and Brooke's laughter is predominantly contingent on the funny moments or jokes shared between them. The laughter pattern generally initiates with Shawn's giggle, leading to Brooke's more pronounced laughter. This pattern showcases the close rapport between the two, characterized by the contagion of laughter.\\
    \bottomrule
    \hline
    \end{tabular}
  }
  \label{tab:selfinst_prompt_and_examples}
  % \vspace{-2mm}
\end{table*}

\subsection{Additional Qualitative Results}\label{sec:appendix_additional_quals}
We provide additional qualitative results on SMILE-Next in \Fref{fig:appendix_textualized_example}-(a).
As illustrated, the proposed laughter-expert LLM produces more accurate and coherent predictions at reasoning tasks compared to the competing models, including audio-visual LLMs and visual LLMs.

\subsection{Example of Textualized Multimodal Cues}
\label{sec:appendix_human_study_example}
We report the example of a textualized multimodal cue at~\Fref{fig:appendix_textualized_example}-(b). As shown, proposed textualized multimodal cues are annotated per video segments, with the transcription, and visual description of the scene, facial action units describing the person's facial expression, and the numeric acoustic features. With such textualized multimodal cues, proposed LLM can analyze the given cues of the video and understand how and why the laughter happened in the context.

\subsection{Human Annotations for SMILE-Next}
We report the instruction and the screenshot of AMT human study at laughter type classification at~\Fref{fig:human_amt_study}. 
For laughter type classification, the annotators were instructed to pick the laughter type with the confidence of the answers, and also they write why they have picked that laughter type in complete sentences. Each sample was annotated by three annotators, and the final laughter type label was determined via majority voting.
For laughter reasoning, the annotators were instructed to determine whether the given expert LLM (\eg, GPT4)'s pseudo-reasoning label is correct. If it is correct, the pseudo-reasoning label is used for the annotation. If not, annotators are instructed to write a free-form answer about why the laughter has occurred, and such samples are used for the laughter reasoning annotations. All payments are adequate considering the participants’ demographics.

\begin{figure*}[t]
    \centering
    \includegraphics[width=0.8\textwidth]{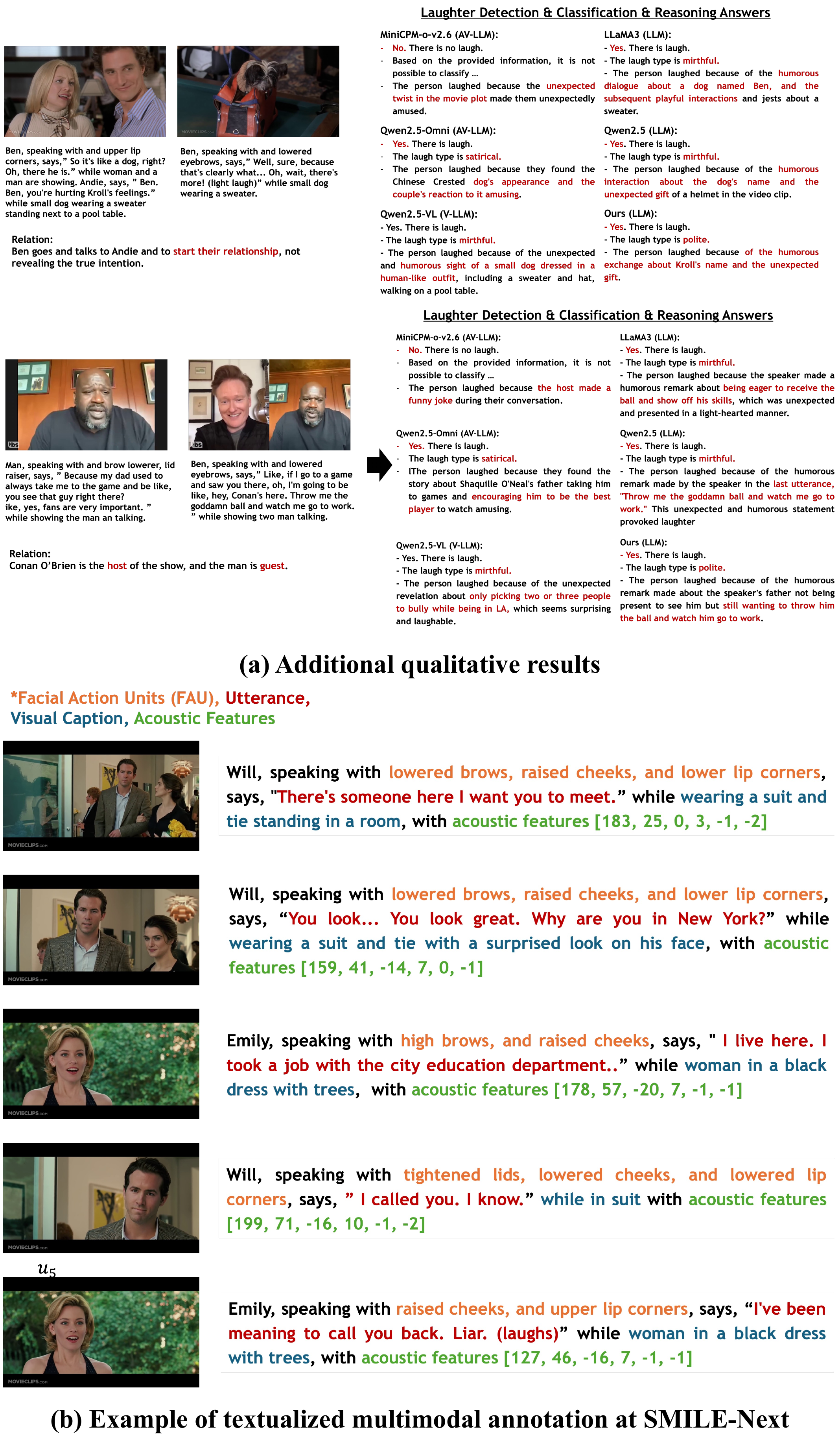}
    \caption{\textbf{Additional qualitative results and example of textualized multimodal cues.} 
    }
    \label{fig:appendix_textualized_example}
\end{figure*}

\begin{figure*}[t]
    \centering
    \includegraphics[width=1\textwidth]{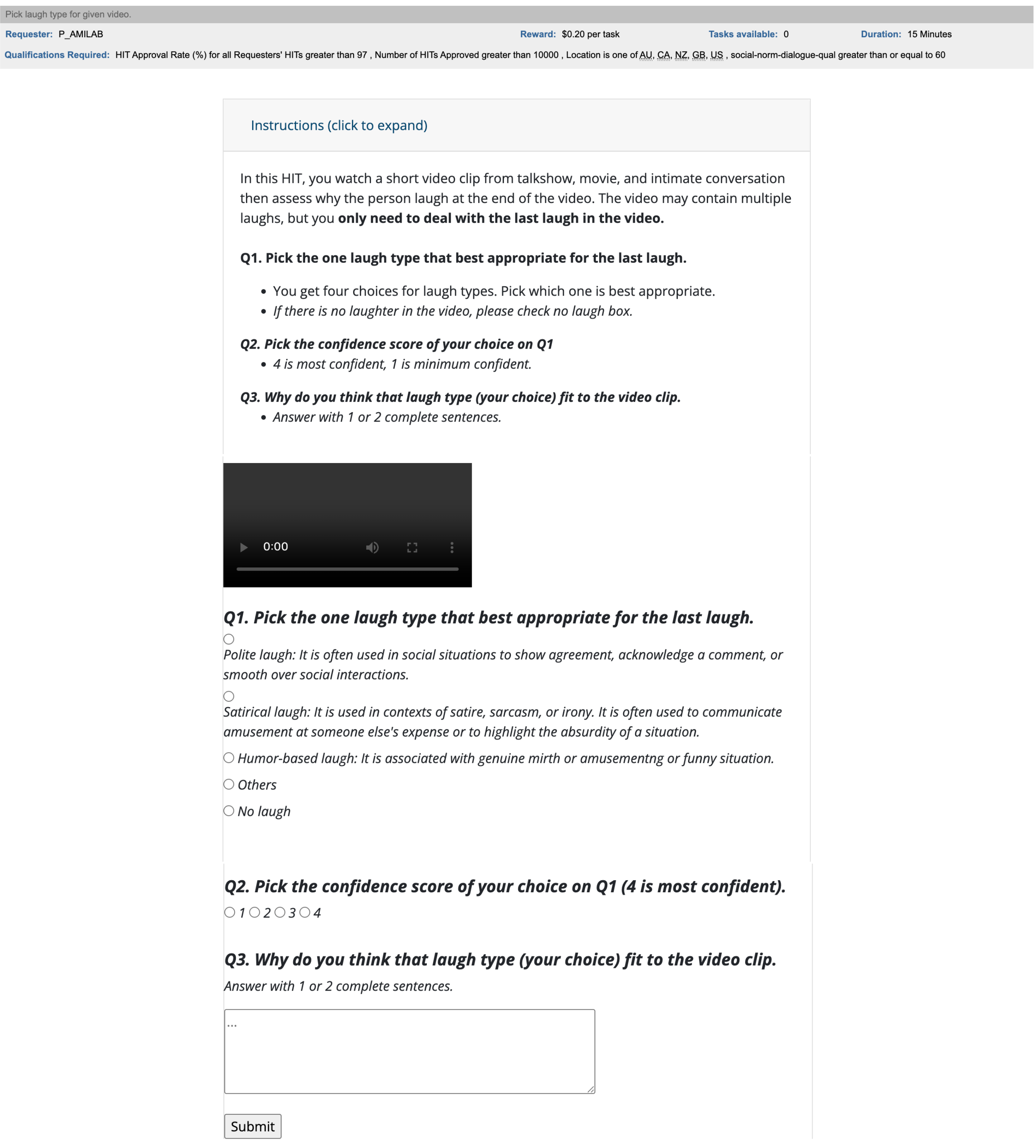}
    \caption{\textbf{Example screenshot of Amazon Mechanical Turk for human annotation. For laughter classification, the human evaluators are instructed to pick a laughter type with a confident, and to write the reason why they choose it.} 
    }
    \label{fig:human_amt_study}
\end{figure*}

\end{document}